\documentclass{article} 
\usepackage{rldmsubmit,palatino}
\usepackage{graphicx}
\usepackage{svg}
\usepackage{natbib}
\usepackage{paralist}
\rldmfinaltrue

\usepackage{amsmath}
\usepackage{mathtools}  
\usepackage{amsfonts}

\usepackage{xspace}
\DeclareRobustCommand{\approach}{\textit{GrowNN}\xspace}
\DeclareRobustCommand{\deeper}{\textit{Net2DeeperNet}\xspace}

\DeclareRobustCommand{\net2net}{\textit{Net2Net}\xspace}

\title{Growing with Experience: Growing Neural Networks in Deep Reinforcement Learning}
\author{
Lukas Fehring \\
Leibniz University Hannover \\
\texttt{l.fehring@ai.uni-hannover.de} \\
\And
Marius Lindauer \\
Leibniz University Hannover \\
\AND
Theresa Eimer \\
Leibniz University Hannover \\
}

\begin{document}
\maketitle
\begin{abstract}
    While increasingly large models have revolutionized much of the machine learning landscape, training even mid-sized networks for Reinforcement Learning (RL) is still proving to be a struggle. This, however, severely limits the complexity of policies we are able to learn. To enable increased network capacity while maintaining network trainability, we propose \approach, a simple yet effective method that utilizes progressive network growth during training. We start training a small network to learn an initial policy. Then we add layers without changing the encoded function. Subsequent updates can utilize the added layers to learn a more expressive policy, adding capacity as the policy's complexity increases. \approach can be seamlessly integrated into most existing RL agents. Our experiments on MiniHack and Mujoco show improved agent performance, with incrementally \approach deeper networks outperforming their respective static counterparts of the same size by up to  $48\%$ on MiniHack Room and $72\%$ on Ant.
\end{abstract}
\section{Introduction}
Scaling deep Reinforcement Learning (RL) agents in terms of network complexity has been shown to be non-trivial~\citep{schwarzer-icml23a, ota-ml24a}. Limiting the size of the function approximator, however, naturally limits the complexity of the resulting policy. Solutions to scaling deep RL include alternative network architectures like mixture of experts~\citep{obandoceron-icml24a} or a combination of different design choices like parameter resets and increased replay ratio with the downside of extending training time~\citep{schwarzer-icml23a}.
This implies a trade-off between using small and large networks for Reinforcement Learning.

Instead, our approach \approach tackles this tradeoff by incrementally growing the network during training, thereby enabling the use of appropriately sized networks, i.e. small networks in the beginning to learn policy fundamentals and larger networks toward the end of training when we need to have the highest policy capacity. To do so, we incrementally increase the number of parameters by growing the network using network morphisms \citep{wei-icml16a}, where additional neurons do not adapt the function encoded in a neural network. We utilize \net2net \citep{chen-iclr16a} transformations to add additional layers and increase the feature extractor depth in PPO~\citep{schulman-arxiv17a}.
This technique, originally proposed to avoid costly retraining when moving from baselines to more complicated models, was successfully used in RL by \cite{berner-arxiv19a} to double the number of policy network parameters, similar to distillation approaches but without an additional imitation learning phase~\citep{wan-automl22a,schwarzer-icml23a}. 
Incremental network growth is a new idea in RL, only employed by \citet{liu-corr24} to increase the network plasticity based on neuron dormancy.

To evaluate the potential of \approach, we conduct experiments on the MiniHack \citep{samvelyan-neurips21} and MuJoCo Ant \citep{todorov-iros12a} environments. \approach enables agents to utilize network depths they otherwise fail with on MiniHack, improving solution rates from $6\%$ to up to $54\%$. Our additional evaluation on Ant shows a relative improvement of up to $72\%$ over networks of unchanging size.
In summary, our contributions are:
\begin{inparaenum}[(i)]
    \item a novel, easy-to-implement growth strategy, \approach, to grow neural networks during agent training and, 
    \item an experimental evaluation showing that our approach allows larger networks to solve previously impossible tasks without algorithmic changes.
\end{inparaenum}

\section{Related Work}
For MDPs with simple observations, the most used network architecture in RL remains the Multi-Layer Perceptron \citep{rumelhart-book85a, parkerholder-jair22a}. For pixel observations, the Nature CNN proposed in the original DQN \citep{mnih-nature15a} and the later proposed Impala CNN \citep{espeholt-icml18a}, i.e., a $15$-layer ResNet \citep{he-cvpr16a}, are often used \citep{parkerholder-jair22a}. All of these architectures are rather small compared to modern vision and foundation models, e.g., the ResNet is usually used with $152$ layers for the ImageNet dataset~\cite{he-cvpr16a}. 
The reason for this development is at least partially that using larger networks for Reinforcement Learning requires design adjustments, and naively scaling up networks with existing algorithms does not tend to work well \citep{schwarzer-icml23a, obandoceron-icml24a, ota-ml24a}. \citet{schwarzer-icml23a} and \citet{ota-ml24a} each combine several changes for training to enable parameter scaling during training, e.g., parameter resets. \citet{obandoceron-icml24a} instead propose to use the Mixture of Experts architecture for Reinforcement Learning. At the same time, several papers propose using transformers \citep{vaswani-neurips17a} for RL, but these are no less difficult to apply to current RL algorithms \citep{Li-tmlr23a}.

Neural Architecture Search (NAS) \citep{elsken-iclr19a} aims to find well-performing architectures automatically but receives little attention within RL \citep{miao-iclr22a}. Nevertheless, methods like DARTS \citep{liu-iclr19a}, initially developed for supervised learning, and BG-PBT \citep{wan-automl22a}, have shown promise when applied to RL, as evidenced by successful implementations in recent work \citep{miao-iclr22a}.
These approaches, however, rely on retraining the network with the new architecture or at least on knowledge distillation from the existing teacher model. 
In contrast, we retain the training progress and extend the capacity for learning from new observations without additional training overhead.

We approach the problem by starting with a small network that grows over time while avoiding retraining from scratch. Similar approaches can be categorized based on how, when, and where the networks are grown \citep{evci-iclr22a}. The \net2net approach we use grows by either adding additional or widening preexisting layers based on intervention. This idea has since been extended, e.g., by \citet{wei-icml16a} and \citet{wen-kdd20a}. In contrast, other approaches propose more complicated, fine-grained operations, often basing their split decisions on the learning behavior, e.g., gradients \citep{wen-kdd20a,mitchell-arxiv24a, pham-ieee24a}. 
\citet{liu-corr24} follow this idea to mitigate the existence of dormant neurons \citep{obandoceron-icml24a} in RL.
While these approaches focus on fine-grained operations that perform many small adjustments to the network's size and instead focus on optimizing its structure, our approach is targeted specifically for increasing parameterization. Thus, it is orthogonal to many of the approaches mentioned above.

\section{Growing the Network}
Learning a well-performing policy in complex environments is an incremental process. To, for example, master controlling eight hinges in MuJoCo's Ant environment~\citep{todorov-iros12a}, the agent needs to learn multiple individual skills, including: (i) how torque applied at each hinge changes the state of all other hinges, (ii) how to stabilize the ant using torque, (iii) how to move forward, and (iv) that large actions are penalized. Importantly, utilizing overly large actions would overshadow the reward signal, and learning to move the ant forward requires first understanding individual leg movement. While small networks may learn the simpler skills easily, putting them together and controlling the ant to move forward is rather complicated, which means a larger network would be preferred. However, a large network could hinder the initial training process.
To tackle this issue, we propose to search for a sequence of networks of appropriate capacity for different points in training.
\approach starts with a small network for the initial training phase and incrementally grows it using the number of environment interactions as a heuristic.

\begin{figure}[ht]
    \vskip 0.2in
    \begin{center}
    \includegraphics[width=0.48\columnwidth]{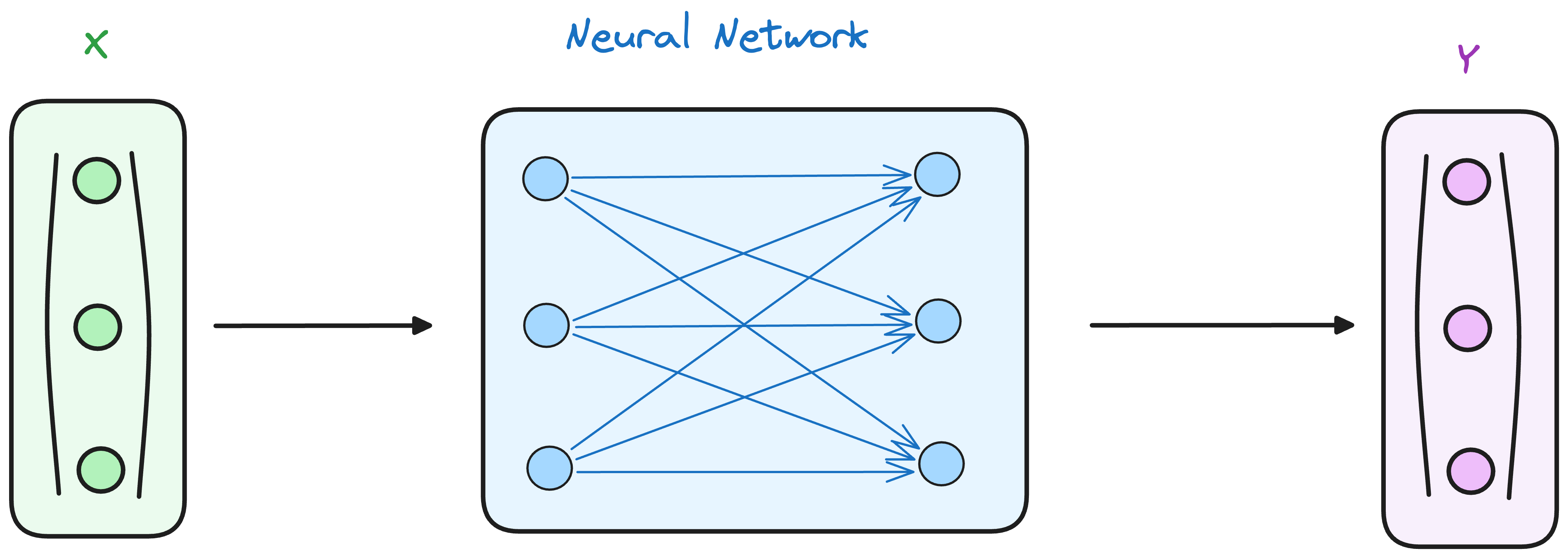}
    \includegraphics[width=0.48\columnwidth]{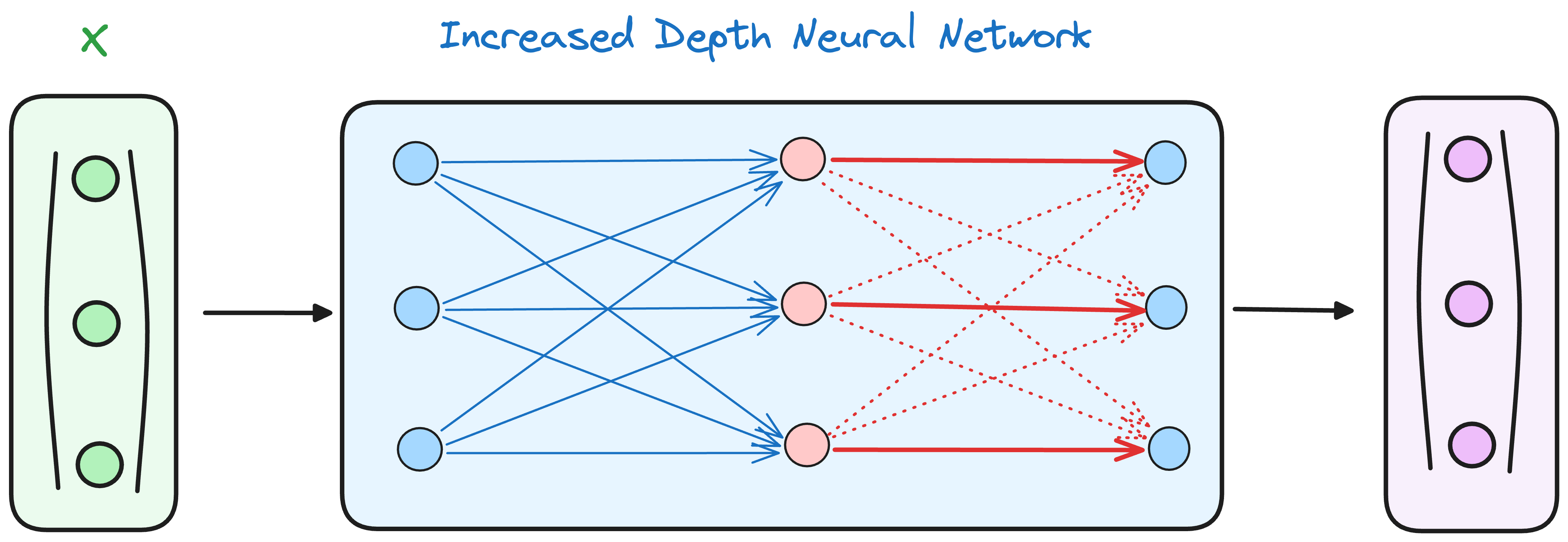}
    \caption{Visualisation of a neural network transformed using \deeper. The added neurons (red circles) are initialized with outgoing weights of either one (solid) for the neuron at the same position or zero (striped) for all other neurons.}
    \label{figure:approach:net2deeper}
    \end{center}
    \vskip -0.2in
\end{figure}

One example of a mechanism enabling this knowledge transfer from a network $\theta_i$ to a new network of increased size $\theta_{i+1}$ is a \emph{Network Morphism}. Formally, a transformation $f: \Theta \to \Theta$ qualifies as a network morphism iff for each input vector $\mathbf{s} \in \mathbf{S}$ the output after transformations is unchanged: ${\theta}(\mathbf{s}) = f({\theta})(\mathbf{s}) \quad \forall \mathbf{s} \in \mathbf{S}$.
This guarantees that the functional mapping remains unchanged despite structural modifications to the network.
We propose using the \net2net transformations proposed by ~\citet{chen-iclr16a}. 
Net2DeeperNet, which we focus on here, adds an identity layer to the network, as shown in Figure \ref{figure:approach:net2deeper}. The transformation maintains the learned policy but adds a new layer. Importantly, to avoid changing the output, the added layer needs to contain no bias, and the activation function needs to be idempotent (such as ReLu) because it is essentially applied twice. 

To reliably train a Reinforcement Learning agent, appropriate hyperparameter configurations need to be selected. 
We optimize them using a modified version of BOHB \citep{falkner-ai18a} with SMAC3~\citep{lindauer-jmlr22a}. Just like in standard BOHB, a set of hyperparameter configurations is sampled, and after a predefined partial training budget, called fidelity, poorly performing runs are terminated. We extend this idea by growing the network after each fidelity. Unlike the original approach, where small fidelities are only considered in the beginning \citep{falkner-ai18a}, we utilize a static fidelity schedule.

\section{Performance Improvements with \textit{\approach} Networks}
\begin{figure}[h!]
    \centering
    \includegraphics[width=0.49\linewidth]{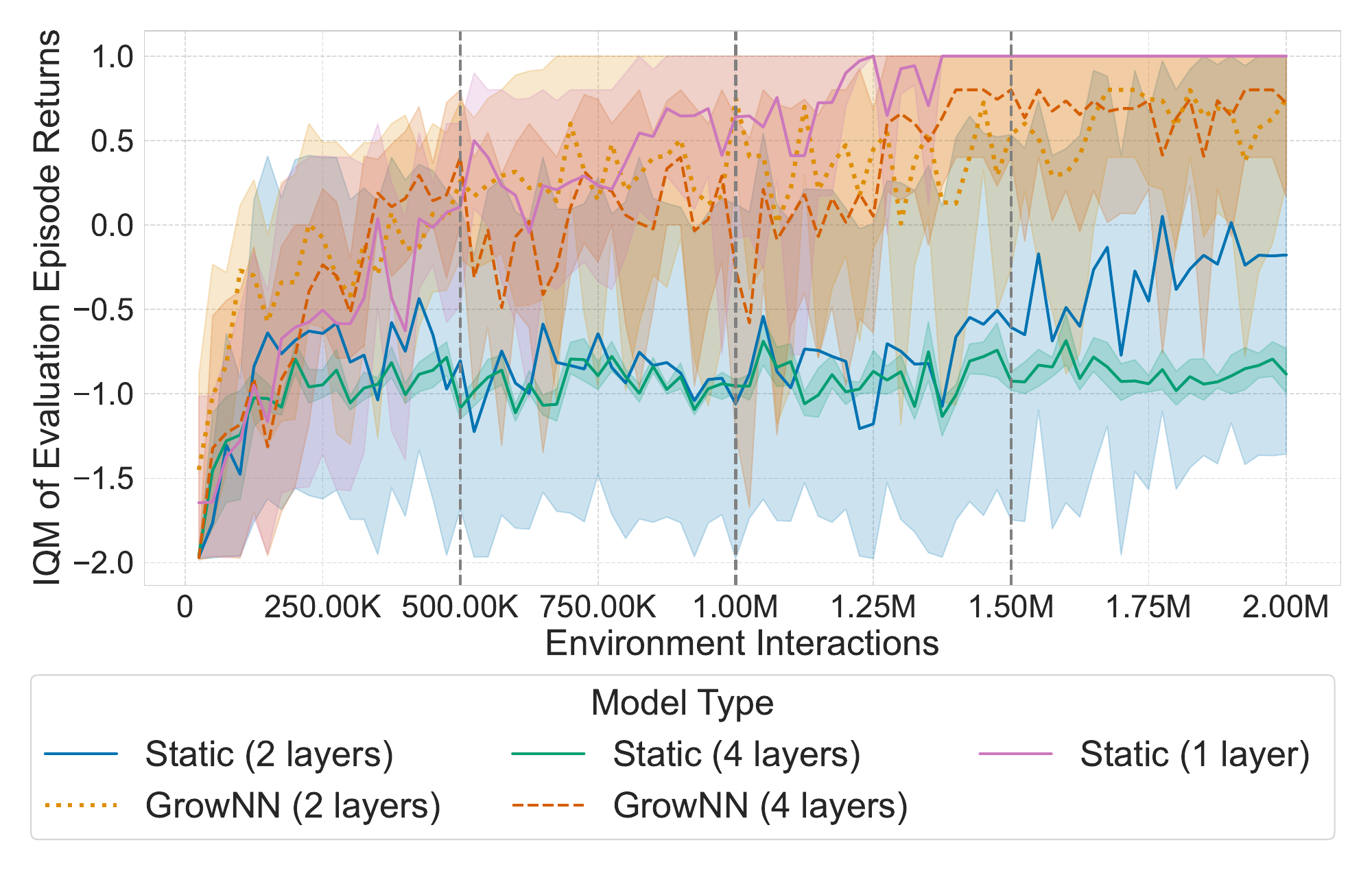}
    \includegraphics[width=0.49\linewidth]{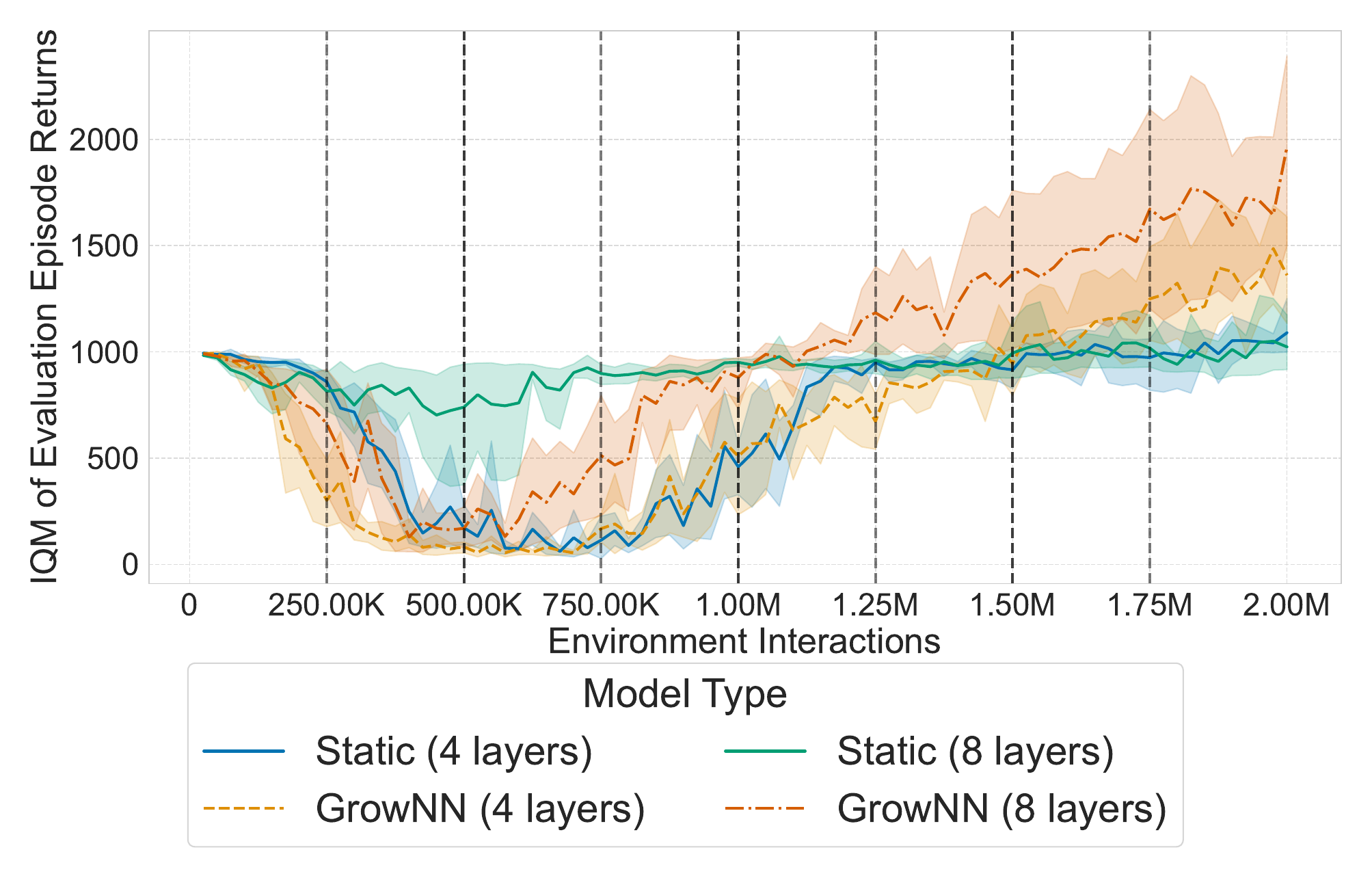}
    \caption{Training curves for static (solid lines) and \approach (dotted lines) networks on MiniHack Room 10x10 (left) and MuJoCo Ant (right). Dotted vertical lines show the evenly spaced growing points for the network. The number of layers for \approach networks refers to the final size.}
    \label{fig:minihack}
\end{figure}
In our experiments, we evaluate the impact of network growth on MiniHack Room \citep{samvelyan-neurips21}, and the MuJoCo Ant \citep{todorov-iros12a} environment.
The agent is trained using the widely-used PPO algorithm \citep{schulman-arxiv17a}.
We use multi-layer perceptrons in all experiments with a single CNN layer for feature encoding on MiniHack. To maximize the impact on the training process, we grow the network in the feature extractor.
As baselines, we use PPO agents equipped with different static network sizes, trained with at least the same budget as the growing agents. Their hyperparameters are tuned with the same approach.

The resulting learning curves for deeper networks can be seen in  Figure \ref{fig:minihack}. 
While on MiniHack, the baseline configured with a static depth $1$ outperforms our growth approach, as soon as we increase the depth of the feature extractor, \approach is superior to statically trained networks with drastically higher solution rates of over $50\%$ compared to $0\%$.
On Ant, the effect is comparable: the baselines do not improve over a return of $1000$, which the environment yields just by avoiding death states. However, the \approach agents learn policies in which the ant actively moves forward with an improvement of $65\%$ in terms of final reward.

\section{Conclusion}
Our initial experiments indicate that the incremental depth increase benefits learning on MiniHack and Ant. 
While more evaluation is necessary, especially on other network architectures like CNNs, we believe growing the network with the agent's capabilities is a promising avenue for scaling RL.
In addition, \approach is independent of algorithm choice and can thus be integrated into most deep RL methods.

\section*{Acknowledgements}
The authors acknowledge funding by the German Research Foundation (DFG): Theresa Eimer under LI 2801/7-1 and Lukas Fehring under LI 2801/10-1.
This work was supported by the Federal Ministry of Education and Research (BMBF), Germany under the AI service center KISSKI (grant no. 01IS22093C) 

\bibliography{bibtex/shortstrings, bibtex/lib,bibtex/my-citations,bibtex/shortproc}
\end{document}